\documentclass[10pt,twocolumn,letterpaper]{article}
\usepackage[pagenumbers]{cvpr}

\usepackage[utf8]{inputenc}
\usepackage[T1]{fontenc}
\usepackage{microtype}
\usepackage{tabularx}
\usepackage{listings}
\usepackage{xurl}
\definecolor{cvprblue}{rgb}{0.21,0.49,0.74}
\usepackage[pagebackref,breaklinks,colorlinks,allcolors=cvprblue]{hyperref}
\newcommand{\venue}[1]{{\itshape\textcolor{gray}{#1}}}
\newcommand{\grayresult}[5]{%
    \textcolor{gray}{\hypersetup{citecolor=gray}#1 \venue{#2}} &
    \textcolor{gray}{#3} & \textcolor{gray}{#4} & \textcolor{gray}{#5} \\
}
\newcommand{\budget}[1]{\rlap{\textsuperscript{#1}}}

\def\confName{CVPR}
\def\confYear{2027}

\title{VidSum-R1: Optimizing Video-to-Video Summarization with GRPO}
\author{Andreas Goulas\textsuperscript{1,2} \quad Evlampios Apostolidis\textsuperscript{1} \quad Vasileios Mezaris\textsuperscript{1} \quad Ioannis Patras\textsuperscript{2}\\
\textsuperscript{1}CERTH-ITI, Thessaloniki, Greece \qquad \textsuperscript{2}Queen Mary University of London, London, UK\\
{\tt\small \{agoulas, apostolid, bmezaris\}@iti.gr, i.patras@qmul.ac.uk}}
\date{}

\begin{document}

\maketitle

\begin{abstract}
Generic video-to-video summarization aims to select informative content from a video to form a video summary under a predefined duration budget. Earlier methods in this domain score frozen visual features or intermediate captions, which limits representation quality. Recent generative approaches based on large language models (LLMs) are typically trained via supervised methods, without explicitly optimizing task-specific metrics. To address these limitations, we introduce VidSum-R1, a framework, which adapts a pretrained video LLM through supervised fine-tuning (SFT) and Group Relative Policy Optimization (GRPO). In contrast to other generative methods that output or score only the selected frames, our model learns an adaptive variable-length interval representation that scores the entire video timeline in a single LLM response. This lets the model adapt the temporal granularity of its scores to each video. SFT establishes this representation from reference importance score annotations. Furthermore, in contrast to prior generative methods that typically perform supervised training only, we introduce GRPO-style training, jointly optimizing two summarization-specific rewards via reinforcement learning (reference-summary overlap and frame-score mean squared error). We evaluate our proposed approach on the large-scale VideoXum and Mr.~HiSum video summarization datasets, showing that our model achieves state-of-the-art results among all compared methods in the literature, across 3 metrics on both datasets, while also studying and highlighting the limitations of the evaluation protocols with respect to the F1 metric.
\end{abstract}

\section{Introduction}

Generic video-to-video summarization helps viewers browse video recordings without viewing the full videos by forming informative video summaries within a predefined duration budget.
This requires judging each moment's importance in the context of the full video.
Whether a segment belongs in the summary depends on the contents of the rest of the video and how much of the budget the remaining content needs.
Unlike script-driven or query-driven summarization, in generic summarization the importance must be inferred from the video alone.

Existing video summarization approaches include scoring networks over frozen visual features \citep{fajtl2019vasnet,apostolidis2021pglsum} or over caption representations from large language models (LLMs) \citep{lee2025llmvs} and generative approaches based on LLMs \citep{hua2025v2xum,ren2024timechat}.
Scoring networks often rely on frozen features with limited representation strength, such as CNN features or captions, while existing generative approaches typically rely on supervised fine-tuning (SFT), which teaches the models to reproduce reference outputs but does not directly optimize summarization-specific quality and fidelity metrics.
In addition, existing generative methods typically output or score only the selected summary frames, making it difficult to calculate video-wide score fidelity metrics.
In the closely related field of video question answering, recent work, such as Video-R1 \citep{feng2025videor1}, has achieved remarkable results by post-training instruction-tuned video LLMs using reinforcement learning (RL), with supervision from verifiable rewards.
For video summarization, optimizing rewards with policy gradients has already been applied to early scoring-based networks \citep{zhou2018drdsn}.
However, to our knowledge, video LLM post-training with RL for video-to-video summarization using summarization-specific rewards has not yet been explored.

\begin{figure*}[t]
    \centering
    \includegraphics[width=\linewidth]{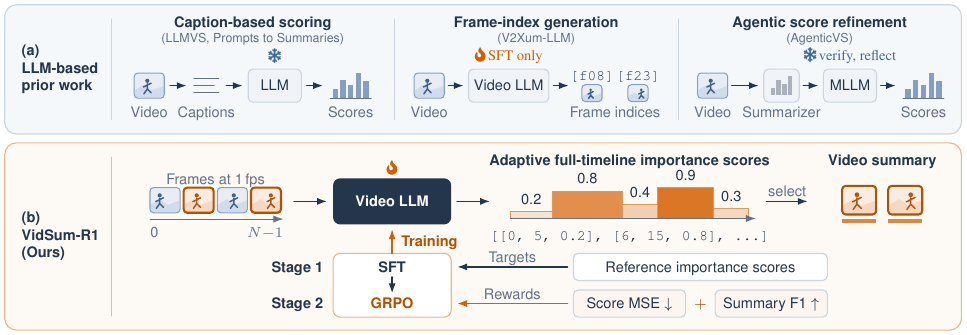}
    \caption{\textbf{VidSum-R1 architecture.} (a) LLM-based prior work: caption-based scoring \citep{lee2025llmvs,barbara2025promptstosummaries}, frame-index generation \citep{hua2025v2xum}, and agentic score refinement \citep{guo2026agenticvs}. (b) In contrast to prior work, VidSum-R1: i) post-trains a video LLM using a two-stage SFT + GRPO framework, jointly optimizing two complementary summarization-specific rewards, and ii) generates importance scores for the entire video timeline through an adaptive variable-length interval representation.}
    \label{fig:overview}
\end{figure*}

To this end, we introduce VidSum-R1, which adapts a pretrained video LLM through SFT followed by Group Relative Policy Optimization (GRPO) \citep{shao2024deepseekmath} with summarization-specific rewards.\footnote{We will make our code publicly available.}
In contrast to earlier generative methods that score only selected frames or segments, our proposed model generates importance scores for the entire video timeline, in a single LLM pass, conditioned on the raw video frames.
This allows us to calculate video-wide metrics of fidelity.
In addition, we use a dense frame importance score representation with variable-length scored intervals, which enables the model to select the temporal granularity of its scores.
Training is performed in two stages:
i) SFT is applied to perform a cold start, establishing the interval-score representation from the reference importance score annotations.
ii) GRPO then explores the large action space of the task, aligning the model outputs jointly with two rewards, namely summary overlap and frame-score accuracy rewards.
Video summarization is particularly well suited to RL-based methods, since we can directly quantify the quality of the generated summaries using such metrics.
This complementary supervision with different metrics is motivated by prior work, such as \citet{otani2019rethinking}, which shows that the evaluation of summaries based on summary-overlap F1 depends on segmentation and selection and proposes additional evaluation metrics.
The overall architecture of VidSum-R1 is shown in \cref{fig:overview}.

We evaluate our model on the large-scale VideoXum \citep{lin2024videoxum} and Mr.~HiSum \citep{sul2023mrhisum} video summarization benchmarks.
Our proposed model achieves the highest F1 summary-overlap score, Kendall's $\tau$ and Spearman's $\rho$ of all compared methods on both benchmarks.
However, we experimentally study the limitations of the original VideoXum protocol, showing that the F1 score should not be used alone when comparing different models (\cref{tab:ablation}).

Our contributions are:
\begin{itemize}
\item \textbf{GRPO post-training for video-to-video summarization.} In contrast to earlier generative methods that mainly rely on supervised training, we propose the use of GRPO for generic video-to-video summarization using video LLMs, jointly optimizing summary overlap and dense frame-score accuracy.
\item \textbf{Adaptive full-timeline importance scoring.} Instead of scoring only the selected frames, VidSum-R1 generates frame importance scores for the entire video timeline, in a single LLM response, conditioned on the raw video frames. We use a dense variable-length scored interval representation, enabling the model to adaptively choose the score granularity per video.
\item \textbf{Study of evaluation protocol limitations.} We show that the original VideoXum protocol retains the frames tied at the selection threshold, yielding summaries beyond the nominal budget and F1 scores that may not be strictly comparable.
We further study this behavior by ablating the selection rule during both GRPO training and evaluation, varying the summary budget, and reporting the percentage of frames that the summaries select.
\end{itemize}

\section{Related work}

\subsection{Video-to-video summarization}

Early methods model temporal dependencies with RNNs \citep{mahasseni2017sumgan,zhou2018drdsn}, interval detection \citep{zhu2021dsnet}, or attention \citep{fajtl2019vasnet,apostolidis2021pglsum}.
DR-DSN \citep{zhou2018drdsn} trains its summarizer with reinforcement learning and a label-free diversity-representativeness reward.
Reinforcement learning has since been used with a video-category classifier reward \citep{zhou2018dqsn}, in an adversarial actor-critic framework \citep{apostolidis2021acsumgan}, and to address exposure bias and non-differentiable metrics \citep{li2018dynseqdpp}.
Other methods model annotator variability with diffusion \citep{kim2025summdiff}, add context from transcripts \citep{he2023a2summ}, audio \citep{badamdorj2021jointva,liu2022umt,joo2026davihd}, or text and audio \citep{kim2026triplesumm}, or align frames with user scripts \citep{mylonas2025sdvsum,mylonas2026sdmvsum}.
Instead of training a summarization network on pre-extracted features, we post-train a pretrained video LLM to predict importance scores.

\subsection{LLM-based video summarization}

Caption-based methods learn frame scores from LLM representations of generated captions (LLMVS \citep{lee2025llmvs}) or convert LLM assessments of scene and video captions into scores (Prompts to Summaries \citep{barbara2025promptstosummaries}).
Instead, our model scores the raw frames directly, retaining visual details that captions may omit.
Among generative methods, V2Xum-LLM \citep{hua2025v2xum} outputs textual frame markers, and the multi-task TimeChat \citep{ren2024timechat} outputs timestamp-score pairs.
HAS \citep{chu2026has} uses a highlight generator to steer LLM attention to important video frames.
AgenticVS \citep{guo2026agenticvs} refines the scores of a frame-scoring summarizer with a multimodal LLM verifier and a reflection step.
Aha \citep{chang2025aha} scores streaming frames with prediction heads on a LoRA-tuned video LLM, and MMDuet \citep{wang2024mmduet} likewise uses a dedicated head, reporting that video LLMs struggle to generate long sequences of relevance scores as text.
These methods rely either on off-the-shelf LLMs or perform supervised fine-tuning, and the generative ones generally output only the selected frames.
In contrast, our video LLM generates scored intervals that cover the whole video, which suits GRPO post-training, since score fidelity can be measured over the entire video.

\subsection{RL post-training of video LLMs}

Following DeepSeekMath \citep{shao2024deepseekmath} and DeepSeek-R1 \citep{guo2025deepseekr1}, GRPO has been adapted to video tasks such as temporal reasoning \citep{feng2025videor1} and grounding \citep{wang2025timer1}.
Closer to our task, TempSamp-R1 \citep{li2025tempsampr1} rewards highlight detection with a recall-weighted timestamp F-measure and a weighted saliency-score error.
MEDitor \citep{ogata2026meditbench} trains Qwen3-VL with GRPO to generate the cuts of message-driven video edits and evaluates zero-shot transfer to summarization on SumMe and TVSum.
Unlike these methods, VidSum-R1 directly optimizes summarization-specific rewards for generic video-to-video summarization, using each benchmark's summary construction protocol.

\subsection{Datasets and evaluation}

Recent large-scale datasets address the data scarcity of the early SumMe \citep{gygli2014summe} and TVSum \citep{song2015tvsum} benchmarks: VideoXum \citep{lin2024videoxum} with human summaries and Mr.~HiSum \citep{sul2023mrhisum} with replay-derived importance scores (\cref{sec:setup}).
SD-VSum \citep{mylonas2025sdvsum} augments VideoXum with scripts, and SD-MVSum \citep{mylonas2026sdmvsum} adds scripts to Mr.~HiSum and transcripts to both.
On SumMe and TVSum, F1 has been shown to depend on the segmentation and selection that turn frame scores into a summary \citep{otani2019rethinking}.
We show that the tie-retaining selection rule of VideoXum can lead to F1 scores that are not directly comparable.

\section{Methodology}
\label{sec:methodology}

\subsection{Interval-score representation}
\label{sec:interval-score}

Our model outputs scored intervals, which lets it select the scoring granularity for each video: segments of uniform importance are scored once, avoiding long sequences of frame scores, which small LLMs may struggle with.
Interval scoring also suits Qwen3-VL, whose visual encoder already merges each pair of adjacent frames into one set of visual tokens \citep{bai2025qwen3vl}.
With the prompt in \cref{fig:prompt}, the model chooses the interval boundaries freely, with no reference segmentation or predefined interval length.

Let $X=(x_0,\ldots,x_{N-1})$ be a video sampled at 1\,fps.
Given $X$, the model generates a JSON text response containing a list of scored intervals $Y=[(a_k,b_k,s_k)]_{k=1}^{K}$.
Here, $a_k\leq b_k$ are the start and end timestamps of each interval (in seconds, inclusive), and $s_k\in[0,1]$ is the interval's importance.
Broadcasting each interval's score to its frames yields $\hat{\mathbf g}\in [0,1]^N$, from which we construct the binary summary $\hat{\mathbf b}\in\{0,1\}^N$ based on the target dataset (\cref{sec:summary-construction}).

Throughout generation, we use regex-constrained decoding \citep{willard2023efficient} to restrict responses to a non-empty list of these triples inside a JSON code fence, with integer timestamps and scores in $[0,1]$ at up to two decimal places, which fixes only the syntax.
We sanitize the response by clipping the intervals to the video, letting later intervals override earlier ones where they overlap, and assigning zero to uncovered frames.
Each video's response token budget accommodates one scored interval per frame, which allows frame-level detail while preventing unbounded decoding.
A response is unparseable if this budget truncates its list or an interval has $a_k>b_k$.

\begin{figure}[htbp]
    \begin{lstlisting}[
        basicstyle=\ttfamily\footnotesize,
        columns=fullflexible,
        breaklines=true,
        breakatwhitespace=true,
        keepspaces=true,
        showstringspaces=false,
        upquote=true,
        frame=single,
        rulecolor=\color{black!40},
        framesep=6pt,
        xleftmargin=7pt,
        xrightmargin=7pt
    ]
Rate how important each part of the video is for a summary of the whole video.
The video is sampled at 1 frame per second starting at 0 seconds and its last frame is at {last_second:g} seconds.
Its timestamps label pairs of consecutive frames by their mean time: <2.5 seconds> precedes the frames at 2 and 3 seconds.
Split the whole video into consecutive time ranges wherever its content or importance changes.
Rate every range with a number from 0.00 (unimportant) to 1.00 (essential) with at most two decimals.
Use the whole scale and reserve the highest scores for the few moments that best summarize the video.
Return ONLY JSON listing the ranges in chronological order as [start_second, end_second, score] with integer seconds and inclusive ends, covering 0 to {last_second:g} seconds:
```json
[[start_second, end_second, score], ...]
```\end{lstlisting}
    \caption{Video summarization prompt.
    }
    \label{fig:prompt}
\end{figure}

\subsection{Summary construction}
\label{sec:summary-construction}

We convert frame scores $\hat{\mathbf g}$ into binary summaries $\hat{\mathbf b}$ following each dataset's reference protocol.
For VideoXum, evaluation follows the original implementation of \citet{lin2024videoxum}, which SD-MVSum \citep{mylonas2026sdmvsum} also adopts.
Every frame scoring at least the $(\lfloor 0.15N\rfloor+1)$-th highest frame score is selected.
Frames tied at this threshold are all retained, so a summary can, and in practice often does, exceed 15\% of the frames.
Unlike SD-MVSum, we do not assign zero F1 to summaries that exceed 25\% of the frames, since this hard cap is not part of the original protocol and makes F1 discontinuous; instead, we report the percentage of selected frames alongside F1.
GRPO rewards use the same rule.

For Mr.~HiSum, GRPO rewards and evaluation both follow its shot-selection protocol \citep{sul2023mrhisum}, a knapsack selection over shots from kernel temporal segmentation \citep{potapov2014kts}, using SD-MVSum's implementation \citep{mylonas2026sdmvsum}.
The knapsack uses mean frame scores as shot values, shot lengths as weights, and a capacity of $\lfloor 0.15N\rfloor$, which need not be filled entirely.
Following the reference Mr.~HiSum implementation, adjacent shots share boundary frames: a shot's value also averages its end boundary, which its length and selected frames exclude.
In addition, following SD-MVSum, the last shot includes the final frame.

\subsection{Supervised fine-tuning}
\label{sec:sft}

Let $\mathbf g\in[0,1]^N$ denote reference frame importance and $\mathbf b^{(m)}\in\{0,1\}^N$ the $m$th of $M$ reference summaries.
For Mr.~HiSum, $\mathbf g$ are the provided replay-derived importance scores, whereas for VideoXum we set $\mathbf g$ to the frame-wise mean of the ten binary reference summaries, so the scores of the two datasets are not directly comparable.
For SFT, we round scores to the nearest multiple of $0.05$ and merge consecutive equal values into intervals covering the full video, so the granularity of the targets follows the reference scores of each video rather than a fixed rate.
We serialize these intervals as JSON triples to form each target response and train with teacher forcing on the negative log-likelihood of the response tokens, normalized by their count in each micro-batch.

\subsection{GRPO rewards and optimization}
\label{sec:grpo}

We use a variant of GRPO \citep{shao2024deepseekmath} with group-centered rewards and token-level loss normalization.
Starting from the SFT model, we sample $G$ responses $z_{ij}$ per video $i\in\mathcal B$ (the videos of an optimizer step) from the current model using regex-constrained decoding at temperature $1$, without top-$k$ or top-$p$ truncation.

For a parseable response $z$, the first reward measures summary overlap by averaging F1, on the $[0,1]$ scale, across reference summaries:
\begin{equation}
    R_{\mathrm{F1}}(z)=\frac{1}{M}\sum_{m=1}^{M}
    \mathrm{F1}\!\left(\hat{\mathbf b},\mathbf b^{(m)}\right).
    \label{eq:f1-reward}
\end{equation}
The second reward is the negative mean squared error between predicted frame scores and (unquantized) reference importance scores across the full timeline, including unselected frames:
\begin{equation}
    R_{\mathrm{MSE}}(z)=-\frac{1}{N}\lVert\hat{\mathbf g}-\mathbf g\rVert_2^2.
    \label{eq:mse-reward}
\end{equation}
The total reward is their weighted sum:
$R(z)=\alpha_{\mathrm{F1}}R_{\mathrm{F1}}(z)+\alpha_{\mathrm{MSE}}R_{\mathrm{MSE}}(z)$.
We use $\alpha_{\mathrm{F1}}=0.5$ and $\alpha_{\mathrm{MSE}}=1$ in all experiments, without searching over them, weighting MSE higher since F1 depends on the scores only through the selected frames, whereas every frame contributes to the MSE term.
Unparseable responses receive a fixed negative reward $R(z)=-1$, which is also the lower bound for parseable responses.

As in Dr.\ GRPO \citep{liu2025drgrpo}, we use group-centered advantages without dividing by the group's standard deviation:
$A_{ij}=R_{ij}-\frac{1}{G}\sum_{j'=1}^{G}R_{ij'}$, where $R_{ij}$ is the reward of $z_{ij}$.
As in DeepSeekMath's experiments \citep{shao2024deepseekmath}, we sample a fresh rollout batch for each optimizer step and use each response once.
Let $Z=\sum_{i\in\mathcal B}\sum_{j=1}^{G}|z_{ij}|$ count the response tokens across all responses in a step, excluding padding.
We use the token-level normalization of DAPO \citep{yu2025dapo}, so the resulting gradient is:
\begin{equation}
    \nabla_\theta\mathcal L_{\mathrm{GRPO}}
    =-\frac{1}{Z}\sum_{i\in\mathcal B}\sum_{j=1}^{G}
    A_{ij}\nabla_\theta\log p_\theta(z_{ij}\mid X_i).
    \label{eq:grpo-gradient}
\end{equation}
Here, $\log p_\theta(z\mid X)$ sums response-token log-probabilities from the model's full-vocabulary softmax (i.e., the regex mask is applied only during sampling).
With each response used once, PPO ratio clipping is inactive, and we use no reference-model KL penalty.

\section{Results}

\subsection{Setup}
\label{sec:setup}

\noindent\textbf{Datasets.}
We evaluate on the large-scale VideoXum and Mr.~HiSum datasets, using the subsets and splits provided by SD-MVSum \citep{mylonas2026sdmvsum}.
\textbf{VideoXum} \citep{lin2024videoxum} contains approximately 14K videos from ActivityNet Captions \citep{krishna2017activitynetcaptions}, each with ten human reference summaries.
\textbf{Mr.~HiSum} \citep{sul2023mrhisum} contains approximately 32K videos selected from YouTube-8M \citep{abuelhaija2016youtube8m}, with replay-derived importance scores and one reference summary per video.
These splits contain 6,782 training and 3,419 test videos for VideoXum, and 26,178 and 1,864 for Mr.~HiSum, as the remaining videos were not publicly available during this work.

\noindent\textbf{Training details.}
We instantiate the video LLM with Qwen3-VL-8B-Instruct \citep{bai2025qwen3vl}.
Video frames keep their native resolution unless the whole video exceeds $8{,}192\times32^2\times 2$ pixels (i.e., about 8,192 visual tokens), in which case all frames are downscaled uniformly, in both training and inference.
We use vLLM \citep{kwon2023efficient} for GRPO rollouts and test inference.
Hyperparameters follow standard practice, without a systematic search.
We train a separate low-rank adapter (LoRA) \citep{hu2022lora} ($r=64$, $\alpha=128$, on the attention and MLP projection layers) for each dataset with the base weights frozen, using AdamW \citep{loshchilov2019adamw} and gradient-norm clipping at $1.0$.
GRPO continues training the SFT adapter, and LoRA dropout ($0.05$) is applied in SFT only.
SFT runs for $4$ epochs on VideoXum and $2$ on the larger Mr.~HiSum, with a batch size of 16, a peak learning rate of $10^{-4}$, linear warmup over 30 and 50 updates, respectively, and cosine decay.
GRPO runs for about one epoch on each dataset (860 updates on VideoXum and 3,440 on Mr.~HiSum), with 8 videos per update, $G=4$ responses per video, a peak learning rate of $5\times10^{-6}$, weight decay $0.1$, linear warmup over 16 updates, and, for Mr.~HiSum only, linear decay over the final 20\% of training.
For both SFT and GRPO, we select the last checkpoint.
As in prior generative works \citep{hua2025v2xum,ren2024timechat}, we report a single training run per dataset.

\noindent\textbf{Evaluation details.}
At test time, we use greedy decoding and the 15\% selection rules of \cref{sec:summary-construction}.
Unparseable responses are evaluated as all-zero frame scores with an empty summary, and constant score vectors count as zero rank correlation, following VideoXum.
We report summary-overlap F1 (\%), averaged across VideoXum's ten references, and Kendall's $\tau$ and Spearman's $\rho$ between predicted frame scores and mean binary annotations (VideoXum) or replay-derived scores (Mr.~HiSum).
All metrics are computed per video and averaged over the test set.

\noindent\textbf{Computational cost.}
All experiments ran on two NVIDIA RTX PRO 6000 Blackwell GPUs.
SFT required approximately 80 GPU hours per dataset, GRPO 48 and 192 GPU hours for VideoXum and Mr.~HiSum, respectively, and each evaluation run about 9 hours on a single GPU.

\subsection{Main results}
\label{sec:mainresults}

\Cref{tab:videoxum-results,tab:mrhisum-results} compare VidSum-R1 with prior work.
In both tables, the PGL-SUM, CSTA, and A2Summ results are taken from SD-MVSum \citep{mylonas2026sdmvsum}.
On VideoXum, VidSum-R1 achieves the highest F1, Kendall's $\tau$, and Spearman's $\rho$ of all compared methods, exceeding the best prior F1, $\tau$, and $\rho$ by 0.2 points, 0.047, and 0.020, respectively.
However, the summary budgets of the compared methods differ (footnotes of \cref{tab:videoxum-results}), ranging from the top 15\% of the frames to about 62\% in the released evaluation code of LGRLN, and some methods use or state no budget.
Only VSUM-BLIP, VTSUM-BLIP, and the results taken from SD-MVSum use the same 15\% budget as VidSum-R1.
The summary length affects F1 even for a random selection, whose F1 rises from 16.2\% at a 15\% budget to 25.5\% at 50\%, whereas the rank correlations do not depend on it.
As the original VideoXum protocol retains the frames tied at the threshold, the summaries of VidSum-R1 cover 41.2\% of the frames on average (\cref{tab:ablation}), so its F1 is not directly comparable with that of shorter summaries, and the rank correlations are the primary comparison.
In \cref{tab:reward-rule}, we compare this selection rule with a stricter one, which never exceeds the budget, in both the GRPO reward and the evaluation.
On Mr.~HiSum, VidSum-R1 also achieves the best result in every metric.
Its rank correlations exceed those of CSTA, the best visual-only method, and of TripleSumm with captions and audio, even though VidSum-R1 observes only the video frames.
The TripleSumm, Joint-VA, and UMT results are taken from \citet{kim2026triplesumm}, who use Inception-v3 visual features (their released data contains 1,909 test videos), and the VASNet and SUM-GAN results from \citet{kim2025summdiff}, who use the same features and the original split.

\noindent\textbf{Ablation study.}
\Cref{tab:ablation} compares VidSum-R1 (SFT + GRPO) with Qwen3-VL without task-specific training (zero-shot) and after SFT only, and reports the mean percentage of selected frames.
On VideoXum, the zero-shot model reaches 28.0\% F1 with near-zero rank correlations, as its tied scores yield summaries that cover 78.7\% of the frames (\cref{sec:summary-construction}), so this F1 is not directly comparable with that of the trained models.
SFT more than triples the correlations but lowers F1, since its training objective exclusively rewards importance score fidelity.
GRPO, which combines the F1 and MSE rewards, improves all three metrics over SFT.
Its summaries, however, select 41.2\% of the frames, compared with 24.4\% for SFT, which we address in the reward-rule ablation below.
On Mr.~HiSum, SFT alone already surpasses the rank correlations of prior visual-only methods (\cref{tab:mrhisum-results}), yet its F1 (49.4\%) is 7.1 points below the uniform-score baseline (\cref{sec:trivial-baseline}).
Only GRPO, which computes its F1 reward through the knapsack selection, exceeds it.
Since the knapsack keeps every summary within the budget and both SFT and SFT + GRPO select 14.7\% of the frames, the 13.0-point F1 gain of GRPO over SFT can be attributed to the improved score predictions and not to summary length.

\noindent\textbf{F1 reward selection rule.}
\Cref{tab:reward-rule} compares GRPO training with the original VideoXum selection rule (\cref{sec:summary-construction}) in the F1 reward against a strict rule, and evaluates both models with the original VideoXum protocol and with the strict rule (F1\textsubscript{strict}).
The strict rule selects up to $\lceil 0.15N\rceil$ highest-scoring positive frames, resolving ties within this count, so that tied scores cannot enlarge the summary and GRPO gains nothing from assigning them.
Under the original evaluation, GRPO with the original rule reaches a 2.5-point higher F1 and higher rank correlations than with the strict rule, but increases the percentage of selected frames from 25.5\% to 41.2\%.
Since the original rule retains all frames tied at the threshold, GRPO tends to learn tied scores that enlarge the summary, which F1 rewards on VideoXum (\cref{sec:trivial-baseline}).
Under the strict evaluation, where no summary exceeds the budget, the order reverses, so each model scores highest under the rule of its own reward.
The F1 advantage of the original rule mainly stems from its longer summaries, but its higher rank correlations do not depend on the selection, so its gains are not due to summary length alone.
Even under the strict evaluation, both models exceed the best prior F1 at a 15\% budget (23.5\%, \cref{tab:videoxum-results}).
With the strict reward rule, the summaries remain close in length to those of SFT, so its 3.9-point F1 gain over SFT under the original evaluation (\cref{tab:ablation}) is not explained by summary length alone.
Our main results use the original rule, so that the reward follows the benchmark's original summary selection.

\noindent\textbf{Statistical significance.}
\Cref{tab:significance} reports the mean per-video improvement of VidSum-R1 over the zero-shot and SFT models on the same test videos, with 95\% paired-bootstrap confidence intervals \citep{efron1993bootstrap} from 100,000 resamples.
All improvements are statistically significant, with confidence intervals well above zero.
Prior methods cannot be tested in this way, as their per-video scores are not available, but on Mr.~HiSum the margins in $\tau$ and $\rho$ over CSTA (0.164 and 0.186), evaluated on the same test split, far exceed the uncertainty in \cref{tab:significance}.
These tests account for the sampling of test videos, not for the variance across training runs.

\begin{table}[!t]
    \centering
    \small
    \setlength{\tabcolsep}{3pt}
    \caption{Video-to-video summarization (V2V-SUM) results on VideoXum. Prior results are as reported in the papers and may use different subsets of the test videos or summary budgets. The gray row uses transcripts. Bold marks the best method and underlining the second best.}
    \label{tab:videoxum-results}
    \begin{tabularx}{\linewidth}{@{}>{\raggedright\arraybackslash}Xrrr@{}}
        \toprule
        Method & F1 (\%) $\uparrow$ & $\tau$ $\uparrow$ & $\rho$ $\uparrow$ \\
        \midrule
        Human \citep{lin2024videoxum} \venue{TMM~24} & 33.8 & 0.305 & 0.336 \\
        Random 15\% & 16.2 & 0.000 & 0.000 \\
        Random 50\% & 25.5 & 0.000 & 0.000 \\
        \midrule
        PGL-SUM \citep{apostolidis2021pglsum} \venue{ISM~21} & 22.1\budget{b} & 0.153 & 0.203 \\
        CSTA \citep{son2024csta} \venue{CVPR~24} & 23.5\budget{b} & 0.176 & 0.233 \\
        VSUM-BLIP \citep{lin2024videoxum} \venue{TMM~24} & 23.1\budget{a} & 0.185 & 0.246 \\
        VTSUM-BLIP \citep{lin2024videoxum} \venue{TMM~24} & 23.5\budget{a} & 0.196 & 0.258 \\
        V2Xum-LLaMA (7B) \citep{hua2025v2xum} \venue{AAAI~25} & 29.0\budget{c} & 0.204 & 0.298 \\
        V2Xum-LLaMA (13B) \citep{hua2025v2xum} \venue{AAAI~25} & 31.6\budget{c} & 0.200 & 0.276 \\
        SpiVG \citep{li2025spivg} \venue{TIP~25} & 31.0\budget{d} & 0.186 & 0.245 \\
        LGRLN \citep{li2026lgrln} \venue{TPAMI~26} & \underline{32.1}\budget{e} & 0.198 & 0.262 \\
        HAS \citep{chu2026has} \venue{ECCV~26} & 32.0\budget{d} & \underline{0.230} & \underline{0.304} \\
        \grayresult{A2Summ \citep{he2023a2summ}}{CVPR~23}{21.3\budget{b}}{0.145}{0.193}
        \midrule
        VidSum-R1 (8B) (Ours) & \textbf{32.3}\budget{a} & \textbf{0.277} & \textbf{0.324} \\
        \bottomrule
    \end{tabularx}
    \par\smallskip
    \begin{minipage}{\linewidth}
        \footnotesize
        Summary budget:
        \textsuperscript{a}top 15\% of the frames, retaining ties at the threshold (\cref{sec:summary-construction}),
        \textsuperscript{b}same as above, but the released code assigns zero F1 to summaries above 25\% of the frames,
        \textsuperscript{c}none (the model generates the summary frames),
        \textsuperscript{d}not stated,
        \textsuperscript{e}the released evaluation code selects the frames scoring above the 37.6th percentile.
    \end{minipage}
\end{table}

\begin{table}[!t]
    \centering
    \small
    \setlength{\tabcolsep}{3pt}
    \caption{Mr.~HiSum results. Prior results are as reported in the papers and may use different visual features, subsets of the test videos, or summary budgets. Gray rows also use transcripts (T), generated captions (C), or audio (A). Bold marks the best result, and underlining the second best among the visual-only methods and among the gray rows.}
    \label{tab:mrhisum-results}
    \begin{tabularx}{\linewidth}{@{}>{\raggedright\arraybackslash}Xrrr@{}}
        \toprule
        Method & F1 (\%) $\uparrow$ & $\tau$ $\uparrow$ & $\rho$ $\uparrow$ \\
        \midrule
        Uniform scores & 56.5 & 0.000 & 0.000 \\
        \midrule
        SUM-GAN \citep{mahasseni2017sumgan} \venue{CVPR~17} & -- & 0.067 & 0.095 \\
        VASNet \citep{fajtl2019vasnet} \venue{ACCVW~18} & -- & 0.069 & 0.102 \\
        PGL-SUM \citep{apostolidis2021pglsum} \venue{ISM~21} & 57.4 & 0.168 & 0.241 \\
        CSTA \citep{son2024csta} \venue{CVPR~24} & \underline{57.7} & \underline{0.193} & \underline{0.274} \\
        SummDiff \citep{kim2025summdiff} \venue{ICCV~25} & -- & 0.175 & 0.238 \\
        TripleSumm (visual) \citep{kim2026triplesumm} \venue{ICLR~26} & -- & 0.187 & 0.258 \\
        \grayresult{Joint-VA\textsuperscript{A} \citep{badamdorj2021jointva}}{ICCV~21}{--}{0.161}{0.231}
        \grayresult{UMT\textsuperscript{A} \citep{liu2022umt}}{CVPR~22}{--}{0.178}{0.253}
        \grayresult{A2Summ\textsuperscript{T} \citep{he2023a2summ}}{CVPR~23}{58.0}{0.169}{0.239}
        \grayresult{HierSum\textsuperscript{T} \citep{beedu2026hiersum}}{CVPRW~26}{58.2}{--}{--}
        \grayresult{DAViHD\textsuperscript{A} \citep{joo2026davihd}}{ICASSP~26}{\underline{59.7}}{0.213}{0.299}
        \grayresult{TripleSumm (full)\textsuperscript{C,A} \citep{kim2026triplesumm}}{ICLR~26}{--}{\underline{0.258}}{\underline{0.352}}
        \midrule
        VidSum-R1 (8B) (Ours) & \textbf{62.4} & \textbf{0.357} & \textbf{0.460} \\
        \bottomrule
    \end{tabularx}
\end{table}

\begin{table}[t]
    \centering
    \small
    \setlength{\tabcolsep}{0pt}
    \caption{Ablation of the training stages, where SFT + GRPO is the full VidSum-R1. F1 is in \%, and Sel.\ is the mean percentage of frames that the summaries select, including ties (with a 15\% budget).}
    \label{tab:ablation}
    \begin{tabular*}{\linewidth}{@{\extracolsep{\fill}}lrrrr>{\hspace{4pt}}rrrr@{}}
        \toprule
        & \multicolumn{4}{c}{VideoXum} & \multicolumn{4}{c}{Mr.~HiSum} \\
        \cmidrule(lr){2-5}\cmidrule(l){6-9}
        Stage & F1 $\uparrow$ & $\tau$ $\uparrow$ & $\rho$ $\uparrow$ & Sel. & F1 $\uparrow$ & $\tau$ $\uparrow$ & $\rho$ $\uparrow$ & Sel. \\
        \midrule
        Zero-shot & 28.0 & 0.054 & 0.061 & 78.7 & 46.7 & 0.025 & 0.031 & 12.8 \\
        SFT & 25.9 & 0.172 & 0.206 & 24.4 & 49.4 & 0.214 & 0.281 & 14.7 \\
        SFT + GRPO & \textbf{32.3} & \textbf{0.277} & \textbf{0.324} & 41.2 & \textbf{62.4} & \textbf{0.357} & \textbf{0.460} & 14.7 \\
        \bottomrule
    \end{tabular*}
\end{table}

\begin{table}[t]
    \centering
    \small
    \setlength{\tabcolsep}{4pt}
    \caption{Ablation of the selection rule in the VideoXum GRPO F1 reward. F1 and Sel.\ follow the original evaluation rule in both rows, and F1\textsubscript{strict} the strict rule (all three in \%).}
    \label{tab:reward-rule}
    \begin{tabular}{@{}lrrrrr@{}}
        \toprule
        Reward rule & F1 $\uparrow$ & $\tau$ $\uparrow$ & $\rho$ $\uparrow$ & Sel. & F1\textsubscript{strict} $\uparrow$ \\
        \midrule
        VidSum-R1 (strict rule) & 29.8 & 0.255 & 0.297 & 25.5 & \textbf{26.4} \\
        VidSum-R1 (original rule) & \textbf{32.3} & \textbf{0.277} & \textbf{0.324} & 41.2 & 24.9 \\
        \bottomrule
    \end{tabular}
\end{table}

\begin{table}[t]
    \centering
    \small
    \setlength{\tabcolsep}{4pt}
    \caption{Mean per-video improvement of VidSum-R1 over the zero-shot (denoted as ZS) and SFT models, with 95\% bootstrap confidence intervals in brackets.
    }
    \label{tab:significance}
    \begin{tabular}{@{}lccc@{}}
        \toprule
        vs. & F1 (\%) & $\tau$ & $\rho$ \\
        \midrule
        \multicolumn{4}{c}{\emph{VideoXum}} \\
        ZS & $4.2\,[3.9,4.5]$ & $.223\,[.216,.232]$ & $.262\,[.254,.271]$ \\
        SFT & $6.4\,[6.2,6.7]$ & $.106\,[.100,.111]$ & $.118\,[.112,.124]$ \\
        \midrule
        \multicolumn{4}{c}{\emph{Mr.~HiSum}} \\
        ZS & $15.7\,[14.5,16.9]$ & $.332\,[.319,.346]$ & $.429\,[.412,.446]$ \\
        SFT & $13.0\,[12.0,14.0]$ & $.143\,[.130,.156]$ & $.179\,[.163,.196]$ \\
        \bottomrule
    \end{tabular}
\end{table}

\begin{figure*}[t]
    \centering
    \includegraphics[width=\linewidth]{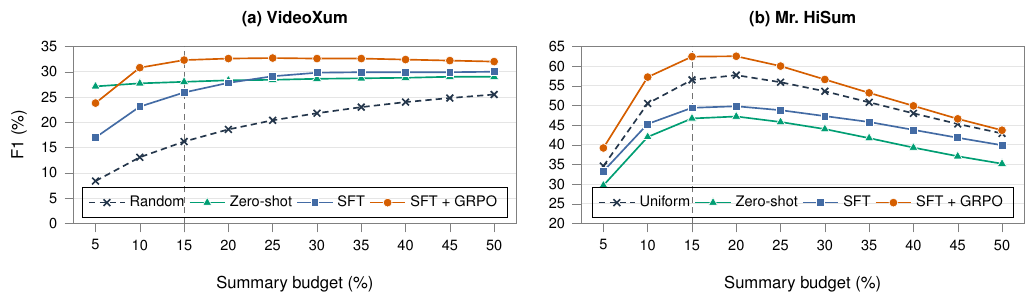}
    \caption{Summary-budget sensitivity. Each panel compares SFT and SFT + GRPO with the zero-shot model and a baseline at budgets from 5\% to 50\% in 5\% increments. The vertical dashed line marks the 15\% GRPO reward budget.}
    \label{fig:budget-sensitivity}
\end{figure*}

\subsection{Uniform and random baselines}
\label{sec:trivial-baseline}

Following \citet{apostolidis2020performance}, who judge F1 relative to a random summarizer, we report a reference baseline for each dataset.
For Mr.~HiSum, assigning uniform frame scores and applying its shot-based summary selection yields 56.5\% F1 (\cref{tab:mrhisum-results}).
This baseline requires no training and observes no video frames, only the provided shot boundaries.
For VideoXum, whose summary selection does not rely on shots, we instead report a random selection baseline (\cref{tab:videoxum-results}): each frame receives an independent uniformly random score, summaries follow the evaluation's selection rule (\cref{sec:summary-construction}) at the given budget, and F1 is averaged over 100 trials per video.
Since the expected precision of random selection is constant across budgets (it equals the reference summary percentage) while its recall grows with the budget, its F1 is highest when all frames are selected, where we calculate it to be 29.2\%.
Under the original rule, constant scores select all frames and thus reach this F1.

\subsection{Summary-budget sensitivity}

In \cref{fig:budget-sensitivity}, we vary the summary budget from 5\% to 50\%, holding predicted scores and reference summaries fixed and retaining each dataset's selection rule.
Rank correlations are unchanged, as only summary selection varies.
On VideoXum, the zero-shot model's F1 is nearly constant, since its mostly tied scores yield summaries that exceed the budget and change little with it (\cref{tab:ablation}).
It thus exceeds SFT at low budgets despite its much lower rank correlations.
SFT + GRPO improves on SFT at every budget on both datasets.
On VideoXum, the gain is largest at 10\% (7.7 points) and decreases to 2.0 points at 50\%, and from the 15\% budget onwards the F1 of SFT + GRPO is nearly constant.
This shows that GRPO maximizes the F1 reward under the given selection rule, often producing tied scores, similarly to the zero-shot model.
Unlike it, however, SFT + GRPO has the highest rank correlations of the three models, and its F1 is the highest at every budget except 5\%, where the zero-shot model exceeds it.
On Mr.~HiSum, the gain is largest at the 15\% GRPO reward budget (13.0 points), and all methods, including the uniform-score baseline, peak at 20\% and decline thereafter, which can be attributed to the knapsack-based selection rather than the predicted scores.
Only SFT + GRPO surpasses the uniform-score baseline.

\subsection{Response quality}

\begin{table}[t]
    \centering
    \small
    \setlength{\tabcolsep}{4pt}
    \caption{Interval granularity and validity, measured from the same outputs used for summary evaluation.}
    \label{tab:interval-analysis}
    \begin{tabular}{@{}lrrrrr@{}}
        \toprule
        & & \multicolumn{3}{c}{Intervals/min} & \\
        \cmidrule(lr){3-5}
        Stage & Tokens & Mean & Std & Median & Valid (\%) \\
        \midrule
        \multicolumn{6}{c}{\emph{VideoXum}} \\
        Zero-shot & 169.9 & 5.9 & 9.4 & 2.5 & 48.6 \\
        SFT & 1467.7 & 51.9 & 5.4 & 52.9 & 99.9 \\
        SFT + GRPO & 1414.9 & 50.2 & 5.4 & 50.8 & 100.0 \\
        \midrule
        \multicolumn{6}{c}{\emph{Mr.~HiSum}} \\
        Zero-shot & 347.6 & 6.9 & 10.5 & 3.5 & 33.2 \\
        SFT & 1213.0 & 26.6 & 7.9 & 24.9 & 100.0 \\
        SFT + GRPO & 1265.2 & 28.1 & 8.8 & 26.6 & 99.8 \\
        \bottomrule
    \end{tabular}
\end{table}

\begin{figure*}[t]
    \centering
    \includegraphics[width=\linewidth]{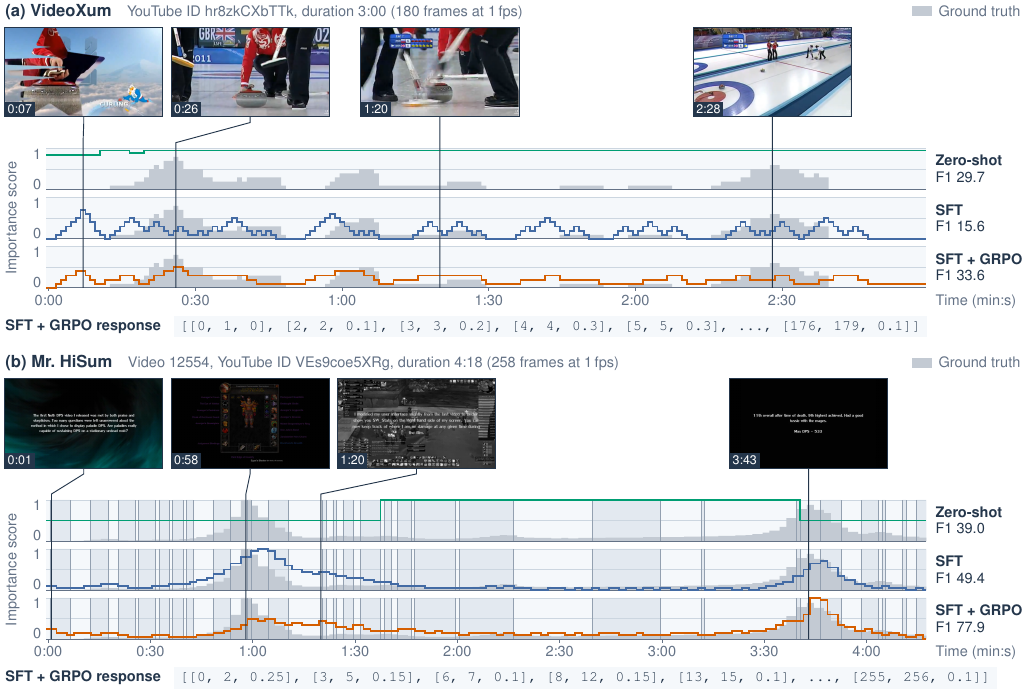}
    \caption{Qualitative examples on VideoXum and Mr.~HiSum comparing SFT and SFT + GRPO with the zero-shot model. Each row plots the frame scores of one model against the ground truth (gray), with the summary F1 (\%) of that model on the video. Dark vertical lines mark the times of the frames shown above, and the last line gives the truncated SFT + GRPO response. On Mr.~HiSum, thin vertical lines and alternating shading mark the shots of the knapsack selection (\cref{sec:summary-construction}).}
    \label{fig:qualitative-examples}
\end{figure*}

\Cref{tab:interval-analysis} reports the mean number of response tokens per video, the intervals-per-minute statistics across videos with parseable responses, and the percentage of all responses with a valid partition of the video.
Validity requires scores in $[0,1]$ and consecutive, non-overlapping intervals with integer boundaries covering $[0,N-1]$ exactly.
With constrained decoding alone, the zero-shot Qwen3-VL model produces a valid partition for only 48.6\% of the VideoXum and 33.2\% of the Mr.~HiSum test videos.
SFT raises validity to almost 100\%, and GRPO preserves it without an explicit reward.
On average, VidSum-R1 generates one interval every 1.2 seconds on VideoXum and every 2.1 seconds on Mr.~HiSum, which is coarser than frame-level scoring.
A fixed rule, such as one interval per frame or per pair of frames merged by the visual encoder, would yield nearly the same rate for every video (60 or 30 intervals per minute), whereas the rate of VidSum-R1 has standard deviations of 5.4 (VideoXum) and 8.8 (Mr.~HiSum) intervals per minute across videos.
GRPO changes this granularity only slightly, so it is largely established by SFT, whose targets follow the reference scores of each video, and it differs between the two datasets.

\subsection{Qualitative examples}
\label{sec:qualitative-examples}

\Cref{fig:qualitative-examples} presents one example from each dataset for the zero-shot, SFT, and SFT + GRPO models.
The zero-shot model assigns coarse, nearly tied scores that do not reflect frame importance. On VideoXum, these ties select most of the frames and thus still yield a higher F1 than SFT.
On VideoXum, SFT scores almost every frame separately, but its scores oscillate throughout the video and only loosely follow the ground truth.
GRPO removes these oscillations: consecutive intervals often share a score, forming plateaus, and its highest scores coincide with the peak of the ground truth.
On Mr.~HiSum, both trained models locate the two highlights, and GRPO follows the ground truth more closely around the second one.
In both examples, GRPO achieves the highest F1.

\section{Conclusions}

VidSum-R1 learns an interval-score representation through SFT and then refines it with GRPO rewards for summary overlap and frame-score accuracy.
The resulting score granularity varies across videos and between the two datasets.
On the large-scale VideoXum and Mr.~HiSum datasets, VidSum-R1 achieves state-of-the-art rank correlations, together with the highest F1 under each benchmark's original summary construction.
We further show that the original VideoXum protocol, which retains tied frames, can lead to F1 scores that are not directly comparable, and that the selection rule of the GRPO reward changes both the F1 and the length of the generated summaries.

\noindent\textbf{Limitations.} Our model relies on visual inputs sampled at 1\,fps, which may miss brief events and omit information conveyed only through audio.
Future work could explore additional task-specific rewards, such as temporal coherence.

\section*{Acknowledgements}

This work was supported by project eXeLMM, implemented in the framework of H.F.R.I. call ``3rd Call for H.F.R.I.'s Research Projects to Support Faculty Members \& Researchers'' (H.F.R.I. Project Number: 25957).

{
    \small

\begin{thebibliography}{51}
\providecommand{\natexlab}[1]{#1}
\providecommand{\url}[1]{\texttt{#1}}
\expandafter\ifx\csname urlstyle\endcsname\relax
  \providecommand{\doi}[1]{doi: #1}\else
  \providecommand{\doi}{doi: \begingroup \urlstyle{rm}\Url}\fi

\bibitem[Abu-El-Haija et~al.(2016)Abu-El-Haija, Kothari, Lee, Natsev, Toderici, Varadarajan, and Vijayanarasimhan]{abuelhaija2016youtube8m}
Sami Abu-El-Haija, Nisarg Kothari, Joonseok Lee, Paul Natsev, George Toderici, Balakrishnan Varadarajan, and Sudheendra Vijayanarasimhan.
\newblock {YouTube-8M}: A large-scale video classification benchmark.
\newblock \emph{arXiv preprint arXiv:1609.08675}, 2016.

\bibitem[Apostolidis et~al.(2020)Apostolidis, Adamantidou, Metsai, Mezaris, and Patras]{apostolidis2020performance}
Evlampios Apostolidis, Eleni Adamantidou, Alexandros~I. Metsai, Vasileios Mezaris, and Ioannis Patras.
\newblock Performance over random: A robust evaluation protocol for video summarization methods.
\newblock In \emph{Proceedings of the 28th ACM International Conference on Multimedia}, pages 1056--1064, 2020.

\bibitem[Apostolidis et~al.(2021{\natexlab{a}})Apostolidis, Adamantidou, Metsai, Mezaris, and Patras]{apostolidis2021acsumgan}
Evlampios Apostolidis, Eleni Adamantidou, Alexandros~I. Metsai, Vasileios Mezaris, and Ioannis Patras.
\newblock {AC-SUM-GAN}: Connecting actor-critic and generative adversarial networks for unsupervised video summarization.
\newblock \emph{IEEE Transactions on Circuits and Systems for Video Technology}, 31\penalty0 (8):\penalty0 3278--3292, 2021{\natexlab{a}}.

\bibitem[Apostolidis et~al.(2021{\natexlab{b}})Apostolidis, Balaouras, Mezaris, and Patras]{apostolidis2021pglsum}
Evlampios Apostolidis, Georgios Balaouras, Vasileios Mezaris, and Ioannis Patras.
\newblock Combining global and local attention with positional encoding for video summarization.
\newblock In \emph{2021 IEEE International Symposium on Multimedia (ISM)}, pages 226--234, 2021{\natexlab{b}}.

\bibitem[Badamdorj et~al.(2021)Badamdorj, Rochan, Wang, and Cheng]{badamdorj2021jointva}
Taivanbat Badamdorj, Mrigank Rochan, Yang Wang, and Li Cheng.
\newblock Joint visual and audio learning for video highlight detection.
\newblock In \emph{Proceedings of the IEEE/CVF International Conference on Computer Vision (ICCV)}, pages 8127--8137, 2021.

\bibitem[Bai et~al.(2025)Bai, Cai, Chen, et~al.]{bai2025qwen3vl}
Shuai Bai, Yuxuan Cai, Ruizhe Chen, et~al.
\newblock {Qwen3-VL} technical report.
\newblock \emph{arXiv preprint arXiv:2511.21631}, 2025.

\bibitem[Barbara and Maalouf(2025)]{barbara2025promptstosummaries}
Mario Barbara and Alaa Maalouf.
\newblock Prompts to summaries: Zero-shot language-guided video summarization with large language and video models.
\newblock \emph{arXiv preprint arXiv:2506.10807}, 2025.

\bibitem[Beedu and Essa(2026)]{beedu2026hiersum}
Apoorva Beedu and Irfan Essa.
\newblock {HierSum}: A global and local attention mechanism for video summarization.
\newblock In \emph{Proceedings of the IEEE/CVF Conference on Computer Vision and Pattern Recognition (CVPR) Workshops}, pages 9069--9079, 2026.

\bibitem[Chang et~al.(2025)Chang, de~Melo, and Lukin]{chang2025aha}
Aiden Chang, Celso de Melo, and Stephanie Lukin.
\newblock {Aha!} -- predicting what matters next: Online highlight detection without looking ahead.
\newblock In \emph{Advances in Neural Information Processing Systems}, pages 8049--8087, 2025.

\bibitem[Chu and Lao(2026)]{chu2026has}
Rui Chu and Yingjie Lao.
\newblock {HAS}: Highlight-guided attention steering for multimodal {LLM} video summarization.
\newblock In \emph{European Conference on Computer Vision (ECCV)}, pages 138--157, 2026.

\bibitem[Efron and Tibshirani(1993)]{efron1993bootstrap}
Bradley Efron and Robert~J. Tibshirani.
\newblock \emph{An Introduction to the Bootstrap}.
\newblock Chapman and Hall/CRC, 1993.

\bibitem[Fajtl et~al.(2019)Fajtl, Sokeh, Argyriou, Monekosso, and Remagnino]{fajtl2019vasnet}
Jiri Fajtl, Hajar~Sadeghi Sokeh, Vasileios Argyriou, Dorothy Monekosso, and Paolo Remagnino.
\newblock Summarizing videos with attention.
\newblock In \emph{Computer Vision -- ACCV 2018 Workshops}, pages 39--54. Springer, 2019.

\bibitem[Feng et~al.(2025)Feng, Gong, Li, Guo, Wang, Peng, Wu, Zhang, Wang, and Yue]{feng2025videor1}
Kaituo Feng, Kaixiong Gong, Bohao Li, Zonghao Guo, Yibing Wang, Tianshuo Peng, Junfei Wu, Xiaoying Zhang, Benyou Wang, and Xiangyu Yue.
\newblock {Video-R1}: Reinforcing video reasoning in {MLLMs}.
\newblock In \emph{Advances in Neural Information Processing Systems}, pages 99114--99137, 2025.

\bibitem[Guo et~al.(2025)Guo, Yang, Zhang, et~al.]{guo2025deepseekr1}
Daya Guo, Dejian Yang, Haowei Zhang, et~al.
\newblock {DeepSeek-R1} incentivizes reasoning in {LLMs} through reinforcement learning.
\newblock \emph{Nature}, 645\penalty0 (8081):\penalty0 633--638, 2025.

\bibitem[Guo et~al.(2026)Guo, Dou, Li, Men, and Jiang]{guo2026agenticvs}
Miaotian Guo, Shuguang Dou, Yin Li, Aidong Men, and Dongsheng Jiang.
\newblock Agentic video summarization via self-reflecting multimodal understanding.
\newblock In \emph{Proceedings of the IEEE/CVF Conference on Computer Vision and Pattern Recognition (CVPR)}, pages 40516--40526, 2026.

\bibitem[Gygli et~al.(2014)Gygli, Grabner, Riemenschneider, and Van~Gool]{gygli2014summe}
Michael Gygli, Helmut Grabner, Hayko Riemenschneider, and Luc Van~Gool.
\newblock Creating summaries from user videos.
\newblock In \emph{Computer Vision -- ECCV 2014}, pages 505--520. Springer, 2014.

\bibitem[He et~al.(2023)He, Wang, Qiu, Bui, Shrivastava, and Wang]{he2023a2summ}
Bo He, Jun Wang, Jielin Qiu, Trung Bui, Abhinav Shrivastava, and Zhaowen Wang.
\newblock Align and attend: Multimodal summarization with dual contrastive losses.
\newblock In \emph{Proceedings of the IEEE/CVF Conference on Computer Vision and Pattern Recognition (CVPR)}, pages 14867--14878, 2023.

\bibitem[Hu et~al.(2022)Hu, Shen, Wallis, Allen-Zhu, Li, Wang, Wang, and Chen]{hu2022lora}
Edward~J. Hu, Yelong Shen, Phillip Wallis, Zeyuan Allen-Zhu, Yuanzhi Li, Shean Wang, Lu Wang, and Weizhu Chen.
\newblock {LoRA}: Low-rank adaptation of large language models.
\newblock In \emph{International Conference on Learning Representations (ICLR)}, 2022.

\bibitem[Hua et~al.(2025)Hua, Tang, Xu, and Luo]{hua2025v2xum}
Hang Hua, Yunlong Tang, Chenliang Xu, and Jiebo Luo.
\newblock {V2Xum-LLM}: Cross-modal video summarization with temporal prompt instruction tuning.
\newblock In \emph{Proceedings of the AAAI Conference on Artificial Intelligence}, pages 3599--3607, 2025.

\bibitem[Joo and Oh(2026)]{joo2026davihd}
Seohyun Joo and Yoori Oh.
\newblock Sounding highlights: Dual-pathway audio encoders for audio-visual video highlight detection.
\newblock In \emph{2026 IEEE International Conference on Acoustics, Speech and Signal Processing (ICASSP)}, pages 13027--13031, 2026.

\bibitem[Kim et~al.(2025)Kim, Hahm, Kim, Sul, Kim, and Lee]{kim2025summdiff}
Kwanseok Kim, Jaehoon Hahm, Sumin Kim, Jinhwan Sul, Byunghak Kim, and Joonseok Lee.
\newblock {SummDiff}: Generative modeling of video summarization with diffusion.
\newblock In \emph{Proceedings of the IEEE/CVF International Conference on Computer Vision (ICCV)}, pages 15096--15106, 2025.

\bibitem[Kim et~al.(2026)Kim, Jeong, Kang, Kim, Oh, and Lee]{kim2026triplesumm}
Sumin Kim, Hyemin Jeong, Mingu Kang, Yejin Kim, Yoori Oh, and Joonseok Lee.
\newblock {TripleSumm}: Adaptive triple-modality fusion for video summarization.
\newblock In \emph{International Conference on Learning Representations (ICLR)}, 2026.

\bibitem[Krishna et~al.(2017)Krishna, Hata, Ren, Fei-Fei, and Niebles]{krishna2017activitynetcaptions}
Ranjay Krishna, Kenji Hata, Frederic Ren, Li Fei-Fei, and Juan~Carlos Niebles.
\newblock Dense-captioning events in videos.
\newblock In \emph{Proceedings of the IEEE International Conference on Computer Vision (ICCV)}, pages 706--715, 2017.

\bibitem[Kwon et~al.(2023)Kwon, Li, Zhuang, Sheng, Zheng, Yu, Gonzalez, Zhang, and Stoica]{kwon2023efficient}
Woosuk Kwon, Zhuohan Li, Siyuan Zhuang, Ying Sheng, Lianmin Zheng, Cody~Hao Yu, Joseph~E. Gonzalez, Hao Zhang, and Ion Stoica.
\newblock Efficient memory management for large language model serving with {PagedAttention}.
\newblock In \emph{Proceedings of the 29th Symposium on Operating Systems Principles}, pages 611--626, 2023.

\bibitem[Lee et~al.(2025)Lee, Gong, and Cho]{lee2025llmvs}
Min~Jung Lee, Dayoung Gong, and Minsu Cho.
\newblock Video summarization with large language models.
\newblock In \emph{Proceedings of the IEEE/CVF Conference on Computer Vision and Pattern Recognition (CVPR)}, pages 18981--18991, 2025.

\bibitem[Li et~al.(2025{\natexlab{a}})Li, Han, Deng, Xiong, and Fan]{li2025spivg}
Wenrui Li, Wei Han, Liang-Jian Deng, Ruiqin Xiong, and Xiaopeng Fan.
\newblock Spiking variational graph representation inference for video summarization.
\newblock \emph{IEEE Transactions on Image Processing}, 34:\penalty0 5697--5709, 2025{\natexlab{a}}.

\bibitem[Li et~al.(2026)Li, Han, Man, Zuo, Fan, and Tian]{li2026lgrln}
Wenrui Li, Wei Han, Hengyu Man, Wangmeng Zuo, Xiaopeng Fan, and Yonghong Tian.
\newblock Language-guided graph representation learning for video summarization.
\newblock \emph{IEEE Transactions on Pattern Analysis and Machine Intelligence}, 48\penalty0 (3):\penalty0 3216--3232, 2026.

\bibitem[Li et~al.(2018)Li, Wang, Yang, and Gong]{li2018dynseqdpp}
Yandong Li, Liqiang Wang, Tianbao Yang, and Boqing Gong.
\newblock How local is the local diversity? reinforcing sequential determinantal point processes with dynamic ground sets for supervised video summarization.
\newblock In \emph{Computer Vision -- ECCV 2018}, pages 156--174. Springer, 2018.

\bibitem[Li et~al.(2025{\natexlab{b}})Li, Cheng, Jia, Kuang, Jiao, Hou, and Cheng]{li2025tempsampr1}
Yunheng Li, Jing Cheng, Shaoyong Jia, Hangyi Kuang, Shaohui Jiao, Qibin Hou, and Ming-Ming Cheng.
\newblock {TempSamp-R1}: Effective temporal sampling with reinforcement fine-tuning for video {LLMs}.
\newblock In \emph{Advances in Neural Information Processing Systems}, pages 40692--40716, 2025{\natexlab{b}}.

\bibitem[Lin et~al.(2024)Lin, Hua, Chen, Li, Hsiao, Ho, and Luo]{lin2024videoxum}
Jingyang Lin, Hang Hua, Ming Chen, Yikang Li, Jenhao Hsiao, Chiuman Ho, and Jiebo Luo.
\newblock {VideoXum}: Cross-modal visual and textural summarization of videos.
\newblock \emph{IEEE Transactions on Multimedia}, 26:\penalty0 5548--5560, 2024.

\bibitem[Liu et~al.(2022)Liu, Li, Wu, Chen, Shan, and Qie]{liu2022umt}
Ye Liu, Siyuan Li, Yang Wu, Chang-Wen Chen, Ying Shan, and Xiaohu Qie.
\newblock {UMT}: Unified multi-modal transformers for joint video moment retrieval and highlight detection.
\newblock In \emph{Proceedings of the IEEE/CVF Conference on Computer Vision and Pattern Recognition (CVPR)}, pages 3042--3051, 2022.

\bibitem[Liu et~al.(2025)Liu, Chen, Li, Qi, Pang, Du, Lee, and Lin]{liu2025drgrpo}
Zichen Liu, Changyu Chen, Wenjun Li, Penghui Qi, Tianyu Pang, Chao Du, Wee~Sun Lee, and Min Lin.
\newblock Understanding {R1-Zero-Like} training: A critical perspective.
\newblock In \emph{Conference on Language Modeling (COLM)}, 2025.

\bibitem[Loshchilov and Hutter(2019)]{loshchilov2019adamw}
Ilya Loshchilov and Frank Hutter.
\newblock Decoupled weight decay regularization.
\newblock In \emph{International Conference on Learning Representations (ICLR)}, 2019.

\bibitem[Mahasseni et~al.(2017)Mahasseni, Lam, and Todorovic]{mahasseni2017sumgan}
Behrooz Mahasseni, Michael Lam, and Sinisa Todorovic.
\newblock Unsupervised video summarization with adversarial {LSTM} networks.
\newblock In \emph{Proceedings of the IEEE Conference on Computer Vision and Pattern Recognition (CVPR)}, 2017.

\bibitem[Mylonas et~al.(2025)Mylonas, Apostolidis, and Mezaris]{mylonas2025sdvsum}
Manolis Mylonas, Evlampios Apostolidis, and Vasileios Mezaris.
\newblock {SD-VSum}: A method and dataset for script-driven video summarization.
\newblock In \emph{Proceedings of the 33rd ACM International Conference on Multimedia}, pages 6596--6604, 2025.

\bibitem[Mylonas et~al.(2026)Mylonas, Zerva, Apostolidis, and Mezaris]{mylonas2026sdmvsum}
Manolis Mylonas, Charalampia Zerva, Evlampios Apostolidis, and Vasileios Mezaris.
\newblock {SD-MVSum}: Script-driven multimodal video summarization method and datasets.
\newblock \emph{IEEE Access}, 14:\penalty0 165224--165242, 2026.

\bibitem[Ogata et~al.(2026)Ogata, Pang, Otani, and Nakashima]{ogata2026meditbench}
Katsuya Ogata, Zongshang Pang, Mayu Otani, and Yuta Nakashima.
\newblock {MEDit-Bench}: A dataset for evaluating message-driven narrative video editing.
\newblock \emph{arXiv preprint arXiv:2607.25300}, 2026.

\bibitem[Otani et~al.(2019)Otani, Nakashima, Rahtu, and Heikkil{\"a}]{otani2019rethinking}
Mayu Otani, Yuta Nakashima, Esa Rahtu, and Janne Heikkil{\"a}.
\newblock Rethinking the evaluation of video summaries.
\newblock In \emph{Proceedings of the IEEE/CVF Conference on Computer Vision and Pattern Recognition (CVPR)}, 2019.

\bibitem[Potapov et~al.(2014)Potapov, Douze, Harchaoui, and Schmid]{potapov2014kts}
Danila Potapov, Matthijs Douze, Zaid Harchaoui, and Cordelia Schmid.
\newblock Category-specific video summarization.
\newblock In \emph{Computer Vision -- ECCV 2014}, pages 540--555. Springer, 2014.

\bibitem[Ren et~al.(2024)Ren, Yao, Li, Sun, and Hou]{ren2024timechat}
Shuhuai Ren, Linli Yao, Shicheng Li, Xu Sun, and Lu Hou.
\newblock {TimeChat}: A time-sensitive multimodal large language model for long video understanding.
\newblock In \emph{Proceedings of the IEEE/CVF Conference on Computer Vision and Pattern Recognition (CVPR)}, pages 14313--14323, 2024.

\bibitem[Shao et~al.(2024)Shao, Wang, Zhu, et~al.]{shao2024deepseekmath}
Zhihong Shao, Peiyi Wang, Qihao Zhu, et~al.
\newblock {DeepSeekMath}: Pushing the limits of mathematical reasoning in open language models.
\newblock \emph{arXiv preprint arXiv:2402.03300}, 2024.

\bibitem[Son et~al.(2024)Son, Park, and Kim]{son2024csta}
Jaewon Son, Jaehun Park, and Kwangsu Kim.
\newblock {CSTA}: Cnn-based spatiotemporal attention for video summarization.
\newblock In \emph{Proceedings of the IEEE/CVF Conference on Computer Vision and Pattern Recognition (CVPR)}, pages 18847--18856, 2024.

\bibitem[Song et~al.(2015)Song, Vallmitjana, Stent, and Jaimes]{song2015tvsum}
Yale Song, Jordi Vallmitjana, Amanda Stent, and Alejandro Jaimes.
\newblock {TVSum}: Summarizing web videos using titles.
\newblock In \emph{Proceedings of the IEEE Conference on Computer Vision and Pattern Recognition (CVPR)}, pages 5179--5187, 2015.

\bibitem[Sul et~al.(2023)Sul, Han, and Lee]{sul2023mrhisum}
Jinhwan Sul, Jihoon Han, and Joonseok Lee.
\newblock {Mr. HiSum}: A large-scale dataset for video highlight detection and summarization.
\newblock In \emph{Advances in Neural Information Processing Systems}, pages 40542--40555, 2023.

\bibitem[Wang et~al.(2024)Wang, Meng, Wang, Liang, Wei, Zhang, and Zhao]{wang2024mmduet}
Yueqian Wang, Xiaojun Meng, Yuxuan Wang, Jianxin Liang, Jiansheng Wei, Huishuai Zhang, and Dongyan Zhao.
\newblock {VideoLLM} knows when to speak: Enhancing time-sensitive video comprehension with video-text duet interaction format.
\newblock \emph{arXiv preprint arXiv:2411.17991}, 2024.

\bibitem[Wang et~al.(2025)Wang, Wang, Xu, et~al.]{wang2025timer1}
Ye Wang, Ziheng Wang, Boshen Xu, et~al.
\newblock {Time-R1}: Post-training large vision language model for temporal video grounding.
\newblock In \emph{Advances in Neural Information Processing Systems}, pages 83330--83364, 2025.

\bibitem[Willard and Louf(2023)]{willard2023efficient}
Brandon~T. Willard and R{\'e}mi Louf.
\newblock Efficient guided generation for large language models.
\newblock \emph{arXiv preprint arXiv:2307.09702}, 2023.

\bibitem[Yu et~al.(2025)Yu, Zhang, Zhu, et~al.]{yu2025dapo}
Qiying Yu, Zheng Zhang, Ruofei Zhu, et~al.
\newblock {DAPO}: An open-source {LLM} reinforcement learning system at scale.
\newblock In \emph{Advances in Neural Information Processing Systems}, pages 113222--113244, 2025.

\bibitem[Zhou et~al.(2018{\natexlab{a}})Zhou, Qiao, and Xiang]{zhou2018drdsn}
Kaiyang Zhou, Yu Qiao, and Tao Xiang.
\newblock Deep reinforcement learning for unsupervised video summarization with diversity-representativeness reward.
\newblock In \emph{Proceedings of the AAAI Conference on Artificial Intelligence}, 2018{\natexlab{a}}.

\bibitem[Zhou et~al.(2018{\natexlab{b}})Zhou, Xiang, and Cavallaro]{zhou2018dqsn}
Kaiyang Zhou, Tao Xiang, and Andrea Cavallaro.
\newblock Video summarisation by classification with deep reinforcement learning.
\newblock In \emph{British Machine Vision Conference (BMVC)}, 2018{\natexlab{b}}.

\bibitem[Zhu et~al.(2021)Zhu, Lu, Li, and Zhou]{zhu2021dsnet}
Wencheng Zhu, Jiwen Lu, Jiahao Li, and Jie Zhou.
\newblock {DSNet}: A flexible detect-to-summarize network for video summarization.
\newblock \emph{IEEE Transactions on Image Processing}, 30:\penalty0 948--962, 2021.

\end{thebibliography}

}

\clearpage
\maketitlesupplementary
\appendix

\section{Environment details}
\label{sec:supp-environment}

\Cref{tab:supp-environment} lists the hardware and software environment of all experiments.

\begin{table}[h]
    \centering
    \small
    \setlength{\tabcolsep}{4pt}
    \caption{Hardware and software environment.}
    \label{tab:supp-environment}
    \begin{tabularx}{\linewidth}{@{}lX@{}}
        \toprule
        Component & Value \\
        \midrule
        GPU & 2 $\times$ NVIDIA RTX PRO 6000 Blackwell (96\,GB) \\
        System memory & 250\,GB \\
        Operating system & Ubuntu 24.04 LTS \\
        CUDA & 12.8 \\
        \midrule
        Python & 3.12.3 \\
        PyTorch & 2.13 \\
        Transformers & 5.16.1 \\
        PEFT & 0.20.0 \\
        vLLM & 0.28.0 \\
        \bottomrule
    \end{tabularx}
\end{table}

\section{GRPO reward ablation}
\label{sec:supp-reward-ablation}

\begin{table}[h]
    \centering
    \small
    \setlength{\tabcolsep}{4pt}
    \caption{Ablation of the GRPO rewards on VideoXum. Sel.\ is the mean percentage of frames that the summaries select, including ties (with a 15\% budget). The first row is the SFT model that all GRPO runs start from, and the last row is the full VidSum-R1.}
    \label{tab:supp-reward-ablation}
    \begin{tabular}{@{}lrrrr@{}}
        \toprule
        GRPO reward & F1 $\uparrow$ & $\tau$ $\uparrow$ & $\rho$ $\uparrow$ & Sel. \\
        \midrule
        None (SFT) & 25.9 & 0.172 & 0.206 & 24.4 \\
        \midrule
        $R_{\mathrm{F1}}$ only & 29.0 & 0.236 & 0.283 & 23.7 \\
        $R_{\mathrm{MSE}}$ only & 29.0 & 0.198 & 0.232 & 39.7 \\
        $R_{\mathrm{F1}}$ + $R_{\mathrm{MSE}}$ & \textbf{32.3} & \textbf{0.277} & \textbf{0.324} & 41.2 \\
        \bottomrule
    \end{tabular}
\end{table}

\Cref{tab:supp-reward-ablation} compares GRPO with each reward of \cref{sec:grpo} alone against the full VidSum-R1, which combines both, on the VideoXum benchmark.
All rows start from the same SFT model and use the same GRPO hyperparameters.
A single-reward run sets the coefficient of the other reward to zero and keeps its own.
Each row is a single training run (\cref{sec:setup}).

Both single-reward runs improve on SFT in every metric, and combining the two rewards improves on both.
The single rewards reach the same F1 in different ways: the F1-only run keeps its summaries as short as those of SFT (23.7\% and 24.4\% of the frames, respectively) and achieves higher rank correlations, whereas the MSE-only run selects a larger percentage of frames.
A likely reason is that the VideoXum reference scores (i.e., the mean of ten binary summaries, see \cref{sec:sft}) take only eleven values and are zero over long ranges (\cref{fig:qualitative-examples}), so predictions that follow them are often tied, and the original rule retains tied frames (\cref{sec:summary-construction}).
Against the MSE-only run, whose summaries have a similar length, the full model gains 3.3 F1 points, so this gain can be attributed to better selections and not to a different summary length.
More importantly, the full VidSum-R1 model outperforms each of the single-reward runs, in terms of the $\tau$ and $\rho$ correlations, by a large margin.

\section{Adaptive and fixed granularity}
\label{sec:supp-granularity}

\Cref{tab:supp-granularity} compares the adaptive interval-score representation of \cref{sec:interval-score} with a fixed-granularity variant, which learns to output fixed-length intervals of 2 seconds.
A fixed interval of 2 seconds directly corresponds to Qwen3-VL's 2-frame aggregation rule.
We compare the two variants under SFT training.
We construct the ground truth responses by quantizing the mean of the reference scores across each 2-second interval and use a separate prompt for this variant.

The adaptive variant shows a clear performance gain over the 2-second fixed variant on VideoXum, according to the $\tau$ and $\rho$ correlations.
This is mainly achieved by learning a more fine-grained representation, on average, compared to the fixed baseline.
Regarding the F1 metric, even though the fixed variant achieves a higher F1 score, this is mainly due to selecting a significantly higher percentage of frames (47.4\% vs 24.4\%).
As explained in \cref{sec:mainresults} of the main paper, a model can reach high F1 scores on VideoXum, without necessarily generating informative predictions.

\begin{table}[t]
    \centering
    \small
    \setlength{\tabcolsep}{4pt}
    \caption{Ablation of the score granularity on VideoXum. Tokens is the mean response length. Models are trained with SFT only.}
    \label{tab:supp-granularity}
    \begin{tabular}{@{}lrrrrr@{}}
        \toprule
        Granularity & F1 $\uparrow$ & $\tau$ $\uparrow$ & $\rho$ $\uparrow$ & Sel. & Tokens \\
        \midrule
        Fixed (2 sec.) & \textbf{26.9} & 0.131 & 0.153 & 47.4 & 837.3 \\
        Adaptive & 25.9 & \textbf{0.172} & \textbf{0.206} & 24.4 & 1467.7 \\
        \bottomrule
    \end{tabular}
\end{table}

\end{document}